\documentclass[11pt,a4paper]{article}
\usepackage[hyperref]{acl2020}
\usepackage{times}
\usepackage{latexsym}

\usepackage{multirow}
\usepackage{latexsym}
\usepackage{mathrsfs}
\usepackage{amsfonts}
\usepackage{amssymb}
\usepackage{amsmath}
\usepackage{bm}
\usepackage{url}
\usepackage{times}
\usepackage{graphicx} 
\usepackage{subfigure}
\usepackage{multicol}
\usepackage{multirow}
\usepackage{booktabs}
\usepackage{colortbl}

\newcommand*\samethanks[1][\value{footnote}]{\footnotemark[#1]}

\newcommand{\bH}{\mathbf{H}}
\newcommand{\bU}{\mathbf{U}}

\usepackage{microtype}

\aclfinalcopy 
\def\aclpaperid{1491} 

\title{Heterogeneous Graph Neural Networks  for  Extractive \\ Document Summarization}

\author{Danqing Wang\thanks{\ \  These two authors contributed equally.}, Pengfei Liu\samethanks, Yining Zheng, Xipeng Qiu\thanks{\ \  Corresponding author.} , Xuanjing Huang\\
  Shanghai Key Laboratory of Intelligent Information Processing, Fudan University \\
  School of Computer Science, Fudan University \\
  825 Zhangheng Road, Shanghai, China \\
\texttt{\{dqwang18,pfliu14,ynzheng19,xpqiu,xjhuang\}@fudan.edu.cn}
}

\date{}

\begin{document}
\maketitle

\begin{abstract}
As a crucial step in extractive document summarization, learning cross-sentence relations has been explored by a plethora of approaches. An intuitive way is to put them
in the graph-based neural network, which has a more complex structure for capturing inter-sentence relationships.
In this paper, we present a \emph{heterogeneous graph-based neural network} for extractive summarization (\textsc{HeterSUMGraph}), which contains semantic nodes of different granularity levels apart from sentences. These additional nodes act as the intermediary between sentences and enrich the cross-sentence relations.
Besides, our graph structure is flexible in natural extension from a single-document setting to multi-document via introducing document nodes.
To our knowledge, we are the first one to introduce different types of nodes into graph-based neural networks for extractive document summarization and perform a comprehensive qualitative analysis to investigate their benefits. The code will be released on Github\footnote{\url{https://github.com/brxx122/HeterSUMGraph}}.

\end{abstract}

\section{Introduction}

Extractive document summarization aims to extract relevant sentences from the original documents and reorganize them as the summary. Recent years have seen a resounding success in the use of deep neural networks on this task \cite{cheng2016neural,narayan2018ranking,arumae2018reinforced,zhong2019searching,liu2019text}.
These existing models mainly follow the encoder-decoder framework in which each sentence will be encoded by neural components with different forms. 

To effectively extract the summary-worthy sentences from a document, a core step is to model the cross-sentence relations. Most current models capture cross-sentence relations with recurrent neural networks (RNNs) \cite{cheng2016neural,nallapati2017summarunner,zhou2018neural}. However, RNNs-based models are usually hard to capture sentence-level long-distance dependency, especially in the case of the long document or multi-documents. One more intuitive way is to model the relations of sentences using the graph structure. Nevertheless, it is challenging to find an effective graph structure for summarization. Efforts have been made in various ways. Early traditional work makes use of inter-sentence cosine similarity to build the connectivity graph like LexRank \cite{erkan2004lexrank} and TextRank \cite{mihalcea2004textrank}.
Recently, some works account for discourse inter-sentential relationships when building summarization graphs, such as the Approximate Discourse Graph (ADG) with sentence personalization features \cite{yasunaga2017graph} and Rhetorical Structure Theory (RST) graph \cite{xu2019discourse}. However, they usually rely on external tools and need to take account of the error propagation problem.
A more straightforward way is to create a sentence-level fully-connected graph. To some extent, the Transformer encoder \cite{vaswani2017attention} used in recent work\cite{zhong2019searching,liu2019text} can be classified into this type, which learns the pairwise interaction between sentences.
Despite their success, how to construct an effective graph structure for summarization remains an open question.




In this paper, we propose a heterogeneous graph network for extractive summarization. Instead of solely building graphs on sentence-level nodes, we introduce more semantic units as additional nodes in the graph to enrich the relationships between sentences. These additional nodes act as the intermediary that connects sentences. Namely, each additional node can be viewed as a special relationship between sentences containing it.
During the massage passing over the heterogeneous graph, these additional nodes will be iteratively updated as well as sentence nodes.

Although more advanced features can be used (e.g., entities or topics), for simplicity, we use words as the semantic units in this paper.
Each sentence is connected to its contained words. There are no direct edges for all the sentence pairs and word pairs.
The constructed heterogeneous \emph{word-sentence} graph has the following advantages:
(a) Different sentences can interact with each other in consideration of the explicit overlapping word information.
(b) The word nodes can also aggregate information from sentences and get updated. Unlike ours, existing models usually keep the words unchanged as the embedding layer.
(c) Different granularities of information can be fully used through multiple message passing processes.
(d) Our heterogeneous graph network is expandable for more types of nodes. For example, we can introduce document nodes for multi-document summarization.


We highlight our contributions as follows:

(1) To our knowledge, we are the first one to construct a heterogeneous graph network for extractive document summarization to model the relations between sentences, which contains not only sentence nodes but also other semantic units. Although we just use word nodes in this paper, more superior semantic units (e.g. entities) can be incorporated.


(2) Our proposed framework is very flexible in extension that can be easily adapt from single-document to multi-document summarization tasks.

(3) 
Our model can outperform all existing competitors on three benchmark datasets without the pre-trained language models\footnote{Since our proposed model is orthogonal to the methods that using pre-trained models, we believe our model can be further boosted by taking the pre-trained models to initialize the node representations, which we reserve for the future.}. Ablation studies and qualitative analysis show the effectiveness of our models.

\section{Related Work}

\paragraph{Extractive Document Summarization}
With the development of neural networks, great progress has been made in extractive document summarization. Most of them focus on the encoder-decoder framework and use recurrent neural networks \cite{cheng2016neural,nallapati2017summarunner,zhou2018neural} or Transformer encoders \cite{zhong2019closer,wang2019exploring} for the sentential encoding. Recently, pre-trained language models are also applied in summarization for contextual word representations \cite{zhong2019searching,liu2019text,xu2019discourse,zhong2020extractive}.

Another intuitive structure for extractive summarization is the graph, which can better utilize the statistical or linguistic information between sentences. Early works focus on document graphs constructed with the content similarity among sentences, like LexRank \cite{erkan2004lexrank} and TextRank \cite{mihalcea2004textrank}. Some recent works aim to incorporate a relational priori into the encoder by graph neural networks (GNNs) \cite{yasunaga2017graph,xu2019discourse}. Methodologically, these works only use one type of nodes, which formulate each document as a homogeneous graph.

\paragraph{Heterogeneous Graph for NLP}
Graph neural networks and their associated learning methods (i.e. message passing \cite{gilmer2017neural}, self-attention \cite{velickovic2017graph}) are originally designed for the homogeneous graph where the whole graph shares the same type of nodes.
However, the graph in the real-world application usually comes with multiple types of nodes \cite{shi2016survey}, namely the heterogeneous graph. To model these structures, recent works have made preliminary exploration.
\citet{tu2019multi} introduced a heterogeneous graph neural network to encode documents, entities and candidates together for multi-hop reading comprehension.
\citet{linmei2019heterogeneous} focused on semi-supervised short text classification and constructed a topic-entity heterogeneous neural graph.


For summarization, \citet{wei2012document} proposes a heterogeneous graph consisting of topic, word and sentence nodes and uses the markov chain model for the iterative update. \citet{wang2019user} modify TextRank for their graph with keywords and sentences and thus put forward HeteroRank. Inspired by the success of the heterogeneous graph-based neural network on other NLP tasks, we introduce it to extractive text summarization to learn a better node representation.





\begin{figure}
  \centerline{
    \includegraphics[width=0.65\linewidth]{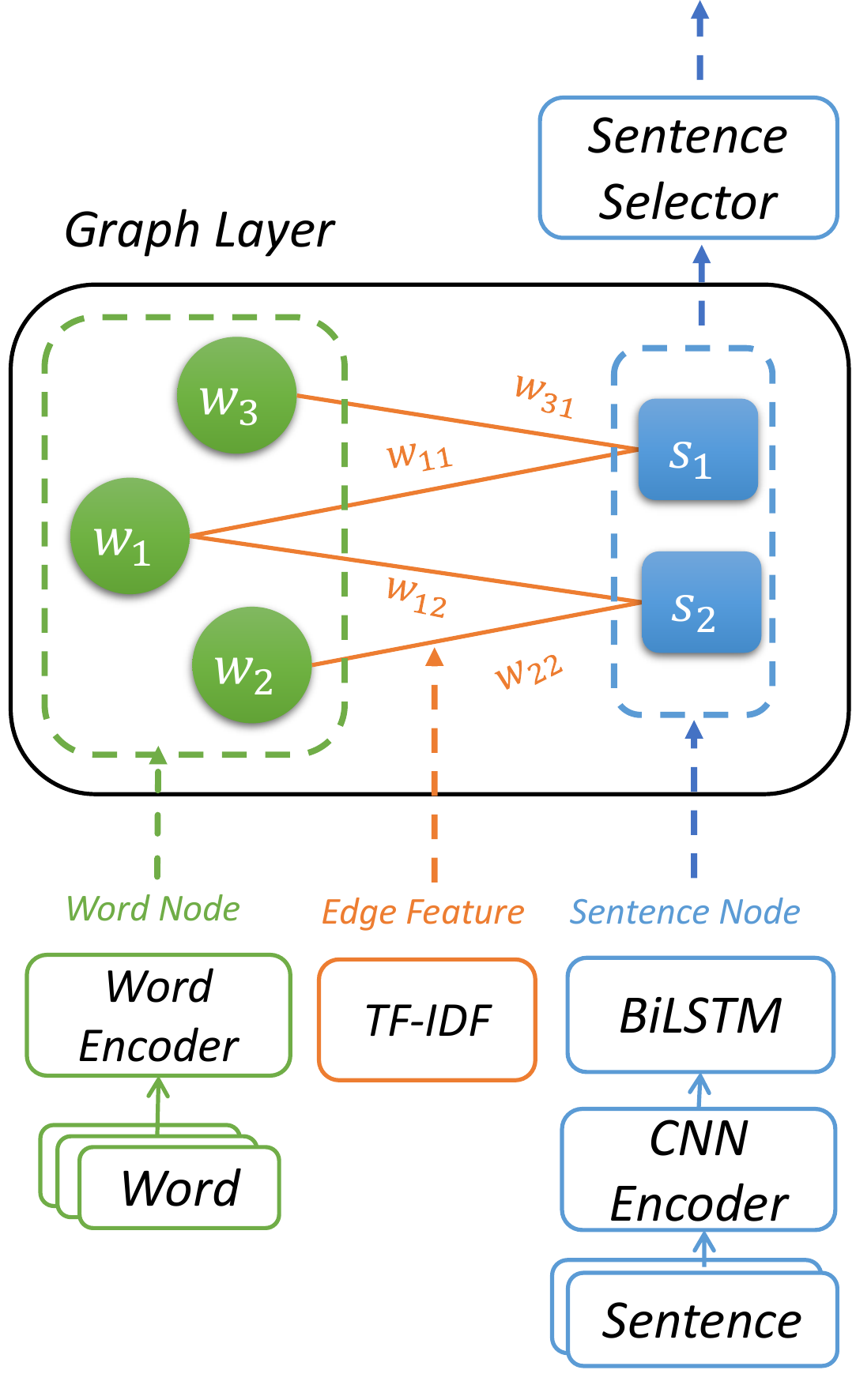}
    }
  \caption{Model Overview. The framework consists of three major modules: \textit{graph initializers}, the \textit{heterogeneous graph layer} and the \textit{sentence selector}. Green circles and blue boxes represent word and sentence nodes respectively. Orange solid lines denote the edge feature (TF-IDF) between word and sentence nodes and the thicknesses indicate the weight. The representations of sentence nodes will be finally used for summary selection.}
  \label{fig:model}
\end{figure}

\section{Methodology}

Given a document $D = \{s_1, \cdots, s_n\}$ with $n$ sentences, we can formulate extractive summarization as a sequence labeling task as \cite{narayan2018ranking, liu2019text}. Our goal is to predict a sequence of labels $y_1, \cdots, y_n$ ($y_i \in \{0,1\}$) for sentences, where $y_i = 1$ represents the $i$-th sentence should be included in the summaries. The ground truth labels, which we call \textsc{oracle}, is extracted using the greedy approach introduced by \citet{nallapati2016abstractive} with the automatic metrics ROUGE \cite{lin2003automatic}.

Generally speaking, our heterogeneous summarization graph consists of two types of nodes: basic semantic nodes (e.g. words, concepts, etc.) as relay nodes and other units of discourse (e.g. phrases, sentences, documents, etc.) as supernodes. Each supernode connects with basic nodes contained in it and takes the importance of the relation as their edge feature. Thus, high-level discourse nodes can establish relationships between each other via basic nodes.

In this paper, we use words as the basic semantic nodes for simplicity. \textsc{HeterSUMGraph} in Section \ref{sec:creation} is a special case which only contains one type of supernodes (sentences) for classification, while \textsc{HeterDocSUMGraph} in Section \ref{sec:multisum} use two (documents and sentences). Based on our framework, other types of supernodes (such as paragraphs) can also be introduced and the only difference lies in the graph structure.


\subsection{Document as a Heterogeneous Graph}
\label{sec:creation}
Given a graph $G = \{V, E\}$,
where $V$ stands for a node set and $E$ represents edges between nodes, our undirected heterogeneous graph can be formally defined as $V=V_w \cup V_s$ and $E=\{e_{11}, \cdots, e_{mn}\}$.
Here, $V_w=\{w_1, \cdots, w_m\}$ denotes $m$ unique words of the document and $V_s=\{s_1, \cdots, s_n\}$ corresponds to the $n$ sentences in the document.
$E$ is a real-value edge weight matrix and $e_{ij} \neq 0$ ($i \in \{1, \cdots, m\}, j \in \{1, \cdots, n\}$) indicates the $j$-th sentence contains the $i$-th word.

Figure \ref{fig:model} presents the overview of our model, which mainly consists of three parts: \textit{graph initializers} for nodes and edges, the \textit{heterogeneous graph layer} and the \textit{sentence selector}. The initializers first create nodes and edges and encode them for the document graph. Then the heterogeneous graph updates these node representations by iteratively passing messages between word and sentence nodes via Graph Attention Network (GAT) \cite{velickovic2017graph}. Finally, the representations of sentence nodes are extracted to predict labels for summaries.

\subsection{Graph Initializers}
\label{sec:init}

Let $\mathbf{X}_w \in \mathbb{R}^{m \times d_w}$ and $\mathbf{X}_s \in \mathbb{R}^{n \times d_s}$ represent the input feature matrix of word and sentence nodes respectively, where $d_w$ is the dimension of the word embedding and $d_s$ is the dimension of each sentence representation vector.
Specifically, we first use Convolutional Neural Networks (CNN) \cite{lecun1998gradient} with different kernel sizes to capture the local n-gram feature for each sentence $l_j$ and then use the bidirectional Long Short-Term Memory (BiLSTM) \cite{hochreiter1997long}  layer to get the sentence-level feature $g_j$. The concatenation of the CNN local feature and the BiLSTM global feature is used as the sentence node feature $X_{s_j}=[l_j ; g_j]$.

To further include information about the importance of relationships between word and sentence nodes, we infuse TF-IDF values in the edge weights. The term frequency (TF) is the number of times $w_i$ occurs in $s_j$ and the inverse document frequency (IDF) is made as the inverse function of the out-degree of $w_i$.

\subsection{Heterogeneous Graph Layer}
\label{sec:graph}
Given a constructed graph $G$ with node features $\textbf{X}_w \cup \textbf{X}_s$ and edge features $\textbf{E}$, we use graph attention networks \cite{velickovic2017graph} to update the representations of our semantic nodes.

We refer to $\bm{h}_i \in \mathbb{R}^{d_h}$, $i \in \{1, \cdots ,(m+n)\}$ as the hidden states of input nodes and the graph attention (GAT) layer  is designed as follows:

\vspace{-1.0em}
{\small
\begin{align}
\label{equ:attnco} z_{i j}&=\mathrm{LeakyReLU} \left(\mathbf{W}_{a}[\mathbf{W}_{q} \bm{h}_{i};\mathbf{W}_{k} \bm{h}_{j}] \right), \\ \label{equ:attn}
\alpha_{i j}&=\frac{\exp (z_{i j})}{\sum_{l \in \mathcal{N}_{i}} \exp (z_{i l})}, \\
\bm{u}_{i} &=\sigma (\sum_{j \in \mathcal{N}_{i}} \alpha_{i j} \mathbf{W}_{v} \bm{h}_{j}),
\end{align}}%
where $\mathbf{W}_{a}$, $\mathbf{W}_{q}$, $\mathbf{W}_{k}$, $\mathbf{W}_{v}$ are trainable weights and $\alpha_{ij}$ is the attention weight between $\bm{h}_i$ and $\bm{h}_j$.
The multi-head attention can be denoted as:
\begin{equation}
\bm{u}_{i}=\Arrowvert_{k=1}^{K} \sigma \left(\sum_{j \in \mathcal{N}_{i}} \alpha_{i j}^{k} \mathbf{W}^{k} \bm{h}_{i} \right).
\end{equation}

Besides, we also add a residual connection to avoid gradient vanishing after several iterations. Therefore, the final output can be represented as:
\begin{equation}
\bm{h}'_{i} = \bm{u}_i + \bm{h}_i.
\end{equation}

We further modify the GAT layer to infuse the scalar edge weights $e_{ij}$, which are mapped to the multi-dimensional embedding space $\bm{e}_{ij} \in \mathbb{R}^{mn \times d_e} $. Thus, Equal \ref{equ:attnco} is modified as follows:
\begin{equation}\small
  z_{i j}=\mathrm{LeakyReLU} \left(\mathbf{W}_{a}[\mathbf{W}_{q} \bm{h}_{i};\mathbf{W}_{k} \bm{h}_{j};\bm{e}_{ij}] \right).
\end{equation}

After each graph attention layer, we introduce a position-wise feed-forward (FFN) layer consisting of two linear transformations just as Transformer \cite{vaswani2017attention}.

\begin{figure}
  \centerline{
    \includegraphics[width=0.8\linewidth]{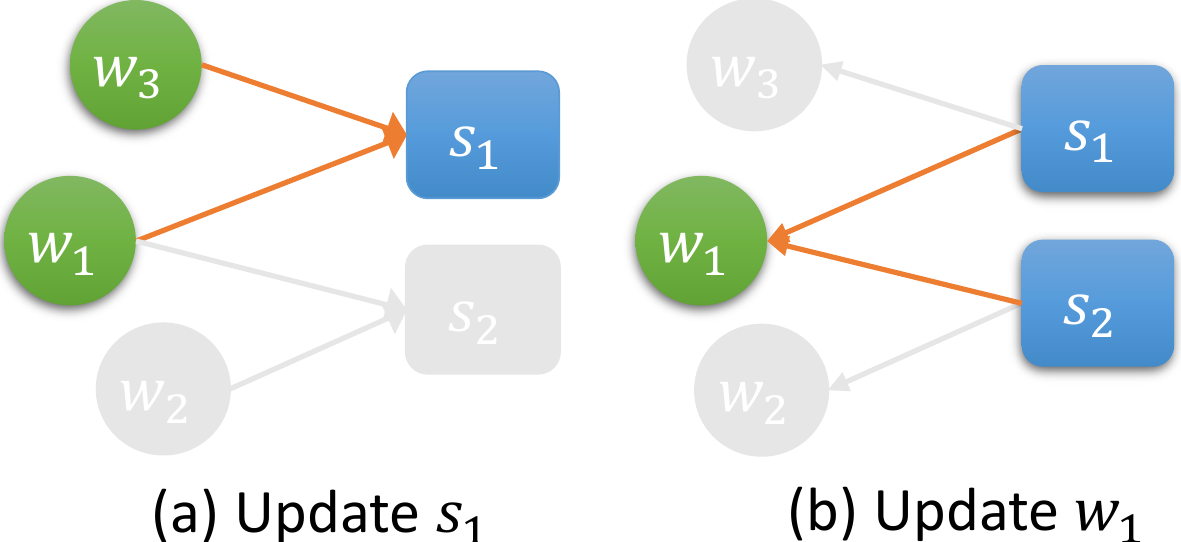}
    }
  \caption{The detailed update process of word and sentence nodes in \textit{Heterogeneous Graph Layer}. Green and blue nodes are word and sentence nodes involved in this turn. Orange edges indicate the current information flow direction. First, for sentence $s_1$, word $w_1$ and $w_3$ are used to aggregate word-level information in (a). Next, $w_1$ is updated by the new representation of $s_1$ and $s_2$ in (b), which are the sentences it occurs. See Section \ref{sec:update} for details on the notation.
  }
  \label{fig:update}
\end{figure}


\paragraph{Iterative updating}
\label{sec:update} To pass messages between word and sentence nodes, we define the information propagation as Figure \ref{fig:update}. Specifically, after the initialization, we update sentence nodes with their neighbor word nodes via the above GAT and FFN layer:

\vspace{-1em}
{\small
\begin{align}
  \bU_{s \leftarrow w}^1 &= \mathrm{GAT}(\bH_s^0, \bH_w^0, \bH_w^0), \\
  \label{equ:update0} \bH_s^1 &= \mathrm{FFN} \left(\bU_{s \leftarrow w}^1 + \bH_s^0 \right),
\end{align}}%
where $\textbf{H}_w^1 = \textbf{H}_w^0 = \textbf{X}_w$, $\textbf{H}_s^0=\textbf{X}_s$ and $\textbf{U}_{s \leftarrow w}^1 \in \mathbb{R}^{m \times d_h}$. $\text {GAT}(\textbf{H}_s^0, \textbf{H}_w^0, \textbf{H}_w^0)$ denotes that $\textbf{H}_s^0$ is used as the attention query and $\textbf{H}_w^0$ is used as the key and value.

After that, we obtain new representations for word nodes using the updated sentence nods and further update sentence nodes iteratively. Each iteration contains a \textit{sentence-to-word} and a \textit{word-to-sentence} update process. For the $t$-th iteration, the process can be represented as:

\vspace{-1em}
{\small
\begin{align}
  \bU_{w \leftarrow s}^{t+1} &= \mathrm {GAT}(\bH_w^t,\bH_s^t,\bH_s^t), \\
  \bH_w^{t+1} &= \mathrm{FFN} \left(\bU_{w \leftarrow s}^{t+1} + \bH_w^t \right), \\
  \bU_{s \leftarrow w}^{t+1} &= \mathrm {GAT}(\bH_s^t,\bH_w^{t+1},\bH_w^{t+1}), \\
  \bH_s^{t+1} &= \mathrm{FFN} \left(\bU_{s \leftarrow w}^{t+1} + \bH_s^t \right).
\end{align}}%

As Figure \ref{fig:update} shows, word nodes can aggregate the document-level information from sentences. For example, the high degree of a word node indicates the word occurs in many sentences and is likely to be the keyword of the document.
Regarding sentence nodes, the one with more important words tends to be selected as the summary.

\subsection{Sentence Selector}
Finally, we need to extract sentence nodes included in the summary from the heterogeneous graph.
Therefore, we do node classification for sentences and cross-entropy loss is used as the training objective for the whole system.

\paragraph{Trigram blocking}
Following \citet{paulus2017deep} and \citet{liu2019text}, we use Trigram Blocking for decoding, which is simple but powerful version of Maximal Marginal Relevance \cite{carbonell1998use}. Specifically, we rank sentences by their scores and discard those which have trigram overlappings with their predecessors.

\begin{figure}
  \vspace{-1em}
  \centerline{
    \includegraphics[width=0.6\linewidth]{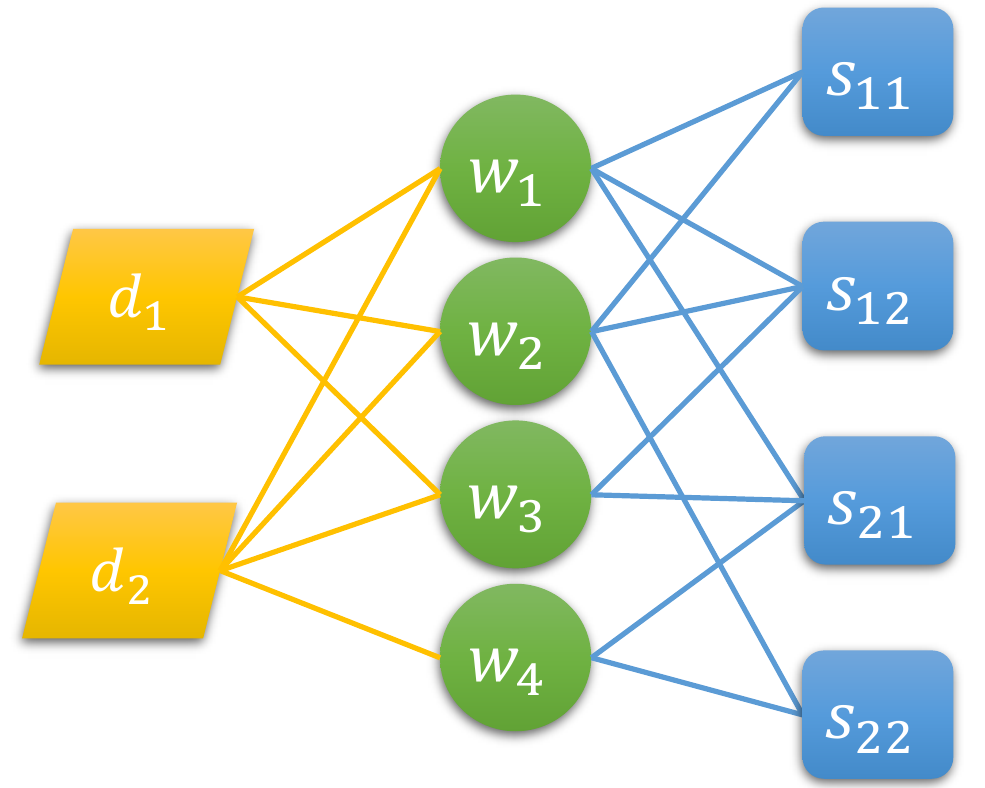}
    }
  \caption{Graph structure of \textsc{HeterDocSUMGraph} for multi-document summarization (corresponding to the \textit{Graph Layer} part of Figure \ref{fig:model}). Green, blue and orange boxes represent word, sentence and document nodes respectively. $d_1$ consists of $s_{11}$ and $s_{12}$ while $d_2$ contains $s_{21}$ and $s_{22}$.
  As a relay node, the relation of \textit{document-document}, \textit{sentence-sentence}, and \textit{sentence-document} can be built through the common word nodes. For example,
    sentence $s_{11}$, $s_{12}$ and $s_{21}$ share the same word $w_1$, which connects them across documents.
  }
  \label{fig:multisum}
\end{figure}

\subsection{Multi-document Summarization}
\label{sec:multisum}
For multi-document summarization, the document-level relation is crucial for better understanding the core topic and most important content of this cluster. However, most existing neural models ignore this hierarchical structure and concatenate documents to a single flat sequence\cite{liu2018generating,fabbri2019multi}. Others try to model this relation by attention-based full-connected graph or take advantage of similarity or discourse relations\cite{liu2019hierarchical}.

Our framework can establish the document-level relationship in the same way as the sentence-level by just adding supernodes for documents(as Figure \ref{fig:multisum}), which means it can be easily adapted from single-document to multi-document summarization.
The heterogeneous graph is then extended to three types of nodes: $V=V_w \cup V_s \cup V_d$ and $V_d=\{d_1, \cdots, d_l\}$ and $l$ is the number of source documents. We name it as \textsc{HeterDocSUMGraph}.

As we can see in Figure \ref{fig:multisum}, word nodes become the bridges between sentences and documents. Sentences containing the same words connect with each other regardless of their distance across documents, while documents establish relationships based on their similar contents.

Document nodes can be viewed as a special type of sentence nodes: a document node connects with contained word nodes and the TF-IDF value is used as the edge weight. Besides, document nodes also share the same update process as sentence nodes. The differences lie in the initialization, where the document node takes the mean-pooling of its sentence node features as its initial state. During the sentence selection, the sentence nodes are concatenated with the corresponding document representations to obtain the final scores for multi-document summarization.


\section{Experiment}
We evaluate our models both on single- and multi-document summarization tasks. Below, we start our experiment with the description of the datasets.

\subsection{Datasets}

\paragraph{CNN/DailyMail}
The CNN/DailyMail question answering dataset \cite{hermann2015teaching,nallapati2016abstractive} is the most widely used benchmark dataset for single-document summarization. The standard dataset split contains 287,227/13,368/11,490 examples for training, validation, and test.
For the data prepossessing, we follow
 \citet{liu2019text}, which use the non-anonymized version as \citet{see2017get}, to get ground-truth labels.

\paragraph{NYT50}
NYT50 is also a single-document summarization dataset, which was collected from New York Times Annotated Corpus \cite{sandhaus2008new} and preprocessed by \citet{durrett2016learning}. It contains 110,540 articles with summaries and is split into 100,834 and 9706 for training and test. Following \citet{durrett2016learning}, we use the last 4,000 examples from the training set as validation and filter test examples to 3,452.

\paragraph{Multi-News}
The Multi-News dataset is a large-scale multi-document summarization introduced by \citet{fabbri2019multi}. It contains 56,216 articles-summary pairs and each example consists of 2-10 source documents and a human-written summary. Following their experimental settings, we split the dataset into 44,972/5,622/5,622 for training, validation and test examples and truncate input articles to 500 tokens.

\subsection{Settings and Hyper-parameters}
For both single-document and multi-document summarization, we limit the vocabulary to 50,000 and initialize tokens with 300-dimensional GloVe embeddings \cite{pennington2014glove}. We filter stop words and punctuations when creating word nodes and truncate the input document to a maximum length of 50 sentences. To get rid of the noisy common words, we further remove 10\% of the vocabulary with low TF-IDF values over the whole dataset. We initialize sentence nodes with $d_s=128$ and edge features $\bm{e}_{ij}$ in $\text{GAT}_e$ with $d_e=50$. Each GAT layer is 8 heads and the hidden size is $d_h=64$, while the inner hidden size of FFN layers is 512.

During training, we use a batch size of 32 and apply Adam optimizer \cite{kingma2014adam} with a learning rate 5e-4. An early stop is performed when valid loss does not descent for three continuous epochs. We select the number of iterations $t=1$ based on the performance on the validation set.\footnote{The detailed experimental results are attached in the Appendix Section.} For decoding, we select top-3 sentences for CNN/DailyMail and NYT50 datasets and top-9 for Multi-New according to the average length of their human-written summaries.

\subsection{Models for Comparison}
\label{sec:models}

\paragraph{Ext-BiLSTM}
Extractive summarizer with BiLSTM encoder learns the cross-sentence relation by regarding a document as a sequence of sentences.
For simplification, we directly take out the initialization of sentence nodes for classification, which includes a CNN encoder for the word level and 2-layer BiLSTM for sentence level. This model can also be viewed as an ablation study of our \textsc{HeterSUMGraph} on the updating of sentence nodes.

\paragraph{Ext-Transformer}
Extractive summarizers with Transformer encoder learn the pairwise interaction \cite{vaswani2017attention} between sentences in a purely data-driven way with a fully connected priori.
Following \cite{liu2019text}, we implement a Transformer-based extractor as a baseline, which contains the same encoder for words followed by 12 Transformer encoder layers for sentences. Ext-Transformer can be regarded as the sentence-level fully connected graph.

\paragraph{\textsc{HeterSUMGraph}}
Our heterogeneous summarization graph model relations between sentences based on their common words, which can be denoted as \textit{sentence-word-sentence} relationships. \textsc{HeterSUMGraph} directly selects sentences for the summary by node classification, while \textsc{HeterSUMGraph} with trigram blocking further utilizes the n-gram blocking to reduce redundancy.


\begin{table}[t]\small\setlength{\tabcolsep}{4pt}
  \centering
  \renewcommand\arraystretch{1}
    \begin{tabular}{lrrr}
    \toprule
    \textbf{Model} & \multicolumn{1}{c}{\textbf{R-1}} & \multicolumn{1}{c}{\textbf{R-2}} & \multicolumn{1}{c}{\textbf{R-L}} \\
    \midrule
    \textsc{Lead-3} \cite{see2017get} & 40.34 & 17.70 & 36.57 \\
    \textsc{Oracle} \cite{liu2019text} & 52.59 & 31.24 & 48.87 \\
    \midrule
    REFRESH \cite{narayan2018ranking} & 40.00 & 18.20 & 36.60 \\
    LATENT \cite{zhang2018neural} & 41.05 & 18.77 & 37.54 \\
    BanditSum \cite{dong2018banditsum}  & 41.50 & 18.70 & 37.60 \\
    NeuSUM \cite{zhou2018neural} & 41.59 & 19.01 & 37.98 \\
    JECS \cite{xu2019neural}  & 41.70 & 18.50 & 37.90 \\
    LSTM+PN \cite{zhong2019searching} & 41.85 & 18.93 & 38.13 \\
    HER w/o Policy \cite{luo2019reading} & 41.70 & 18.30 & 37.10 \\
    HER w Policy \cite{luo2019reading}   & 42.30 & 18.90 & 37.60 \\
    \midrule
    Ext-BiLSTM & 41.59	& 19.03 & 38.04 \\
    Ext-Transformer & 41.33 & 18.83 & 37.65 \\
    \textsc{HSG} & 42.31 & 19.51 & 38.74 \\
    \textsc{HSG} + Tri-Blocking & \textbf{42.95} & \textbf{19.76} & \textbf{39.23} \\
    \bottomrule
    \end{tabular}%
    \caption{Performance (Rouge) of our proposed models against recently released summarization systems on CNN/DailyMail.}
  \label{tab:cnndm}%
\end{table}%

\section{Results and Analysis}
\subsection{Single-document Summarization}
We evaluate our single-document model on CNN/DailyMail and NYT50 and report the unigram, bigram and longest common subsequence overlap with reference summaries by R-1, R-2 and R-L. Due to the limited computational resource, we don't apply pre-trained contextualized encoder (i.e. BERT \cite{devlin2018bert}) to our models, which we will regard as our future work. Therefore, here, we only compare with models without BERT for the sake of fairness.

\paragraph{Results on CNN/DailyMail} Table \ref{tab:cnndm} shows the results on CNN/DailyMail. The first part is the \textsc{Lead-3} baseline and \textsc{oracle} upper bound, while the second part includes other summarization models.

We present our models (described in Section \ref{sec:models}) in the third part. Compared with Ext-BiLSTM, our heterogeneous graphs achieve more than 0.6/0.51/0.7 improvements on R-1, R-2 and R-L, which indicates the cross-sentence relationships learned by our \textit{sentence-word-sentence} structure is more powerful than the sequential structure. Besides, Our models also outperform Ext-Transformer based on fully connected relationships. This demonstrates that our graph structures effectively prune unnecessary connections between sentences and thus improve the performance of sentence node classification.

Compared with the second block of Figure \ref{tab:cnndm}, we observe that \textsc{HeterSUMGraph} outperforms all previous non-BERT-based summarization systems and trigram blocking leads to a great improvement on all ROUGE metrics. Among them, HER \cite{luo2019reading} is a comparable competitor to our \textsc{HeterSUMGraph}, which formulated the extractive summarization task as a contextual-bandit problem and solved it with reinforcement learning. Since the reinforcement learning and our trigram blocking plays a similar role in reorganizing sentences into a summary \cite{zhong2019searching}, we additionally compare HER without policy gradient with \textsc{HeterSUMGraph}. Our \textsc{HeterSUMGraph} achieve 0.61 improvements on R-1 over HER without policy for sentence scoring, and \textsc{HeterSUMGraph} with trigram blocking outperforms by 0.65 over HER for the reorganized summaries.



\renewcommand\arraystretch{1.3}
\begin{table}[htbp]\small\setlength{\tabcolsep}{4pt}
  \centering  \renewcommand\arraystretch{1}
    \begin{tabular}{lrrr}
    \toprule
    \textbf{Model} & \multicolumn{1}{c}{\textbf{R-1}} & \multicolumn{1}{c}{\textbf{R-2}} & \multicolumn{1}{c}{\textbf{R-L}} \\
    \midrule
    First sentence \cite{durrett2016learning}  & 28.60 & 17.30 & - \\
    First k words \cite{durrett2016learning}  & 35.70 & 21.60 & - \\
    \textsc{Lead-3}  & 38.99 & 18.74 & 35.35 \\
    \textsc{Oracle} & 60.54 & 40.75 & 57.22 \\
    \midrule
    COMPRESS \cite{durrett2016learning} & 42.20 & 24.90 & - \\
    SUMO \cite{liu2019single}  & 42.30 & 22.70 & 38.60 \\
    PG* \cite{see2017get} & 43.71 & 26.40 &  - \\
    DRM \cite{paulus2017deep}   & 42.94 & 26.02 & - \\
    \midrule
    Ext-BiLSTM & 46.32 & 25.84 & 42.16 \\
    Ext-Transformer & 45.07 & 24.72 & 40.85 \\
    \textsc{HSG}  & \textbf{46.89} & \textbf{26.26} & \textbf{42.58} \\
    \textsc{HSG} + Tri-Blocking & 46.57 & 25.94 & 42.25 \\
    \bottomrule
    \end{tabular}%
    \caption{Limited-length ROUGE Recall on NYT50 test set. The results of models with * are copied from \citet{liu2019text} and '-' means that the original paper did not report the result.}
  \label{tab:nyt50}%
\end{table}%

\paragraph{Results on NYT50} Results on NYT50 are summarized in Table \ref{tab:nyt50}. Note that we use limited-length ROUGE recall as \citet{durrett2016learning}, where the selected sentences are truncated to the length of the human-written summaries and the recall scores are used instead of F1. The first two lines are baselines given by \citet{durrett2016learning} and the next two lines are our baselines for extractive summarization. The second and third part report the performance of other non-BERT-based works and our models respectively.

Again, we observe that our cross-sentence relationship modeling performs better than BiLSTM and Transformer.
Our models also have strong advantages over other non-BERT-based approaches on NYT50.
Meanwhile, we find trigram block doesn't work as well as shown on CNN/DailyMail, and we attribute the reason to the special formation of summaries of CNN/DailyMail dataset.
\footnote{
\citet{nallapati2016abstractive} concatenate summary bullets, which are written for different parts of the article and have few overlaps with each other, as a multi-sentence summary. However, when human write summaries for the whole article (such as NYT50 and Multi-News), they will use key phrases repeatedly. This means roughly removing sentences by n-gram overlaps will lead to loss of important information.}

\paragraph{Ablation on CNN/DailyMail} In order to better understand the contribution of different modules to the performance, we conduct ablation study using our proposed \textsc{HeterSUMGraph} model on CNN/DailyMail dataset. First, we remove the filtering mechanism for low TF-IDF words and the edge weights respectively.  We also remove residual connections between GAT layers. As a compensation, we concatenate the initial sentence feature after updating messages from nearby word nodes in Equal \ref{equ:update0}:
\begin{align}
  \bH_s^1 &= \mathrm{FFN} \left([\bU_{s \leftarrow w}^1 ; \bH_s^0] \right).
\end{align}
Furthermore, we make iteration number $t=0$, which deletes the word updating and use the sentence representation $\textbf{H}_s^1$ for classification. Finally, we remove the BiLSTM layer in the initialization of sentence nodes.

As Table \ref{tab:ablation} shows, the removal of low TF-IDF words leads to increases on R-1 and R-L but drops on R-2. We suspect that filtering noisy words enable the model to better focus on useful word nodes, at the cost of losing some bigram information. The residual connection plays an important role in the combination of the original representation and the updating message from another type of nodes, which cannot be replaced by the concatenation. Besides, the introduction of edge features, word update and BiLSTM initialization for sentences also show their effectiveness.

\begin{table}[htbp]\small
  \centering  \renewcommand\arraystretch{1}
    \begin{tabular}{lrrr}
    \toprule
    \textbf{Model} & \multicolumn{1}{c}{\textbf{R-1}} & \multicolumn{1}{c}{\textbf{R-2}} & \multicolumn{1}{c}{\textbf{R-L}} \\
    \midrule
    \textsc{HSG} & 42.31 & 19.51 & 38.74 \\
     - filter words & 42.24 & 19.56 & 38.68 \\
     - edge feature & 42.14 & 19.41 & 38.60 \\
     - residual connection & 41.59 & 19.08 & 38.05 \\
     - sentence update & 41.59 & 19.03 & 38.04 \\
     - word update & 41.70 & 19.16 & 38.15 \\
     - BiLSTM & 41.70 & 19.09 & 38.13 \\
    \bottomrule
    \end{tabular}%
    \caption{Ablation studies on CNN/DailyMail test set. We remove various modules and explore their influence on our model. '-' means we remove the module from the original \textsc{HeterSUMGraph}. Note that \textsc{HeterSUMGraph} without the updating of sentence nodes is actually the Ext-BiLSTM model described in Section \ref{sec:models}.}
  \label{tab:ablation}%
\end{table}%

\subsection{Multi-document Summarization}
We first take the concatenation of the First-k sentences from each source document as the baseline and use the codes and model outputs\footnote{https://github.com/Alex-Fabbri/
Multi-News} released by \citet{fabbri2019multi} for other models.

To explore the adaptability of our model to multi-document summarization, we concatenate multi-source documents to a single mega-document and apply \textsc{HeterSUMGraph} as the baseline. For comparison, we extend \textsc{HeterSUMGraph} to multi-document settings \textsc{HeterDocSUMGraph} as described in Section \ref{sec:multisum}. Our results are presented in Table \ref{tab:multinews}.

Specifically, we observe that both of our \textsc{HeterSUMGraph} and \textsc{HeterDocSUMGraph} outperform previous methods while \textsc{HeterDocSUMGraph} achieves better performance improvements. This demonstrates the introduction of document nodes can better model the document-document relationships and is beneficial for multi-document summarization.
As mentioned above, trigram blocking does not work for the Multi-News dataset, since summaries are written as a whole instead of the concatenations of summary bullets for each source document.



\begin{table}[htbp]\footnotesize\setlength{\tabcolsep}{1pt}
  \centering  \renewcommand\arraystretch{1}
    \begin{tabular}{lrrr}
    \toprule
    \textbf{Model} & \multicolumn{1}{c}{\textbf{R-1}} & \multicolumn{1}{c}{\textbf{R-2}} & \multicolumn{1}{c}{\textbf{R-L}} \\
    \midrule
    First-1  & 25.44 & 7.06  & 22.12 \\
    First-2  & 35.70 & 10.28 & 31.71 \\
    First-3  & 40.21 & 12.13 & 37.13 \\
    \textsc{Oracle} & 52.32 & 22.23 & 47.93 \\
    \midrule
    LexRank* \cite{erkan2004lexrank}  & 41.77 & 13.81 & 37.87 \\
    TextRank* \cite{mihalcea2004textrank} & 41.95 & 13.86 & 38.07 \\
    MMR* \cite{carbonell1998use}   & 44.72 & 14.92 & 40.77 \\
    PG$\dag$ \cite{lebanoff2018adapting} & 44.55 & 15.54 & 40.75 \\
    BottomUp$^\dag$ \cite{gehrmann2018bottom}  & 45.27 & 15.32 & 41.38 \\
    Hi-MAP$^\dag$ \cite{fabbri2019multi} & 45.21 & 16.29 & 41.39 \\
    \midrule
    \textsc{HSG} & 45.66 & 16.22 & 41.80 \\
    \textsc{HSG} + Tri-Blocking & 44.92 & 15.59 & 40.89 \\
    \textsc{HDSG} & \textbf{46.05} & \textbf{16.35} & \textbf{42.08} \\
    \textsc{HDSG} + Tri-Blocking & 45.55 & 15.78 & 41.29 \\
    \bottomrule
    \end{tabular}%
    \caption{Results on the test set of Multi-News. We reproduce models with `*' via the released code and directly use the outputs of $\dag$ provided by \citet{fabbri2019multi} for evaluation. }
  \label{tab:multinews}%
\end{table}%

\subsection{Qualitative Analysis}
We further design several experiments to probe into how our \textsc{HeterSUMGraph} and \textsc{HeterDocSUMGraph} help the single- and multi-document summarization.

\paragraph{Degree of word nodes}
In \textsc{HeterSUMGraph}, the degree of a word node indicates its occurrence across sentences and thus can measure the redundancy of the document to some extent. Meanwhile, words with a high degree can aggregate information from multiple sentences, which means that they can benefit more from the iteration process. Therefore, it is important to explore the influence of the node degree of words on the summarization performance.

\begin{figure}
  \centerline{
    \includegraphics[width=1\linewidth]{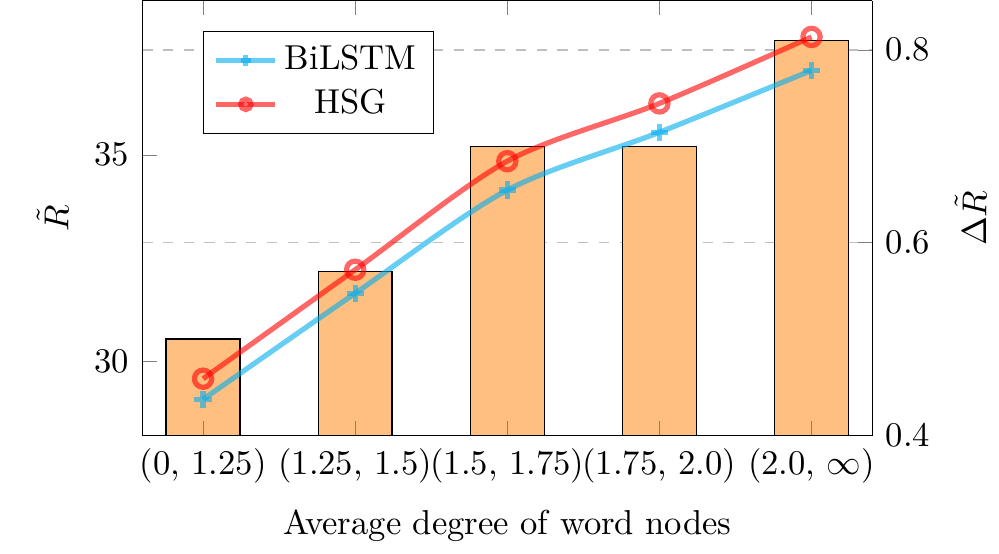}
    }
  \caption{Relationships between the average degree of word nodes of the document (x-axis) and $\Tilde{R}$, which is the mean of R-1, R-2 and R-L (lines for left y-axis), and between $\Delta \Tilde{R}$, which is the delta $\Tilde{R}$ of \textsc{HeterSUMGraph} and Ext-BiLSTM (histograms for right y-axis).}
  \label{fig:degree}
\end{figure}

We first calculate the average degree of word nodes for each example based on the constructed graph. Then the test set of CNN/DailyMail is divided into 5 intervals based on it (x-axis in Figure \ref{fig:degree}). We evaluate the performance of \textsc{HeterSUMGraph} and Ext-BiLSTM in various parts and the mean score of R-1, R-2, R-L is drawn as lines (left y-axis $\Tilde{R}$). The ROUGE increases with the increasing of the average degree of word nodes in the document, which means that articles with a high redundancy are easier for neural models to summarize.

To make $\Delta \Tilde{R}$ between models more obvious, we draw it with histograms (right y-axis). From Figure \ref{fig:degree}, we can observe that \textsc{HeterSUMGraph} performs much better for documents with a higher average word node degree. This proves that the benefit brought by word nodes lies in the aggregation of information from sentences and the propagation of their global representations.

\paragraph{Number of source documents}
We also investigate how the number of source documents influences the performance of our model. To this end, we divide the test set of Multi-News into different parts by the number of source documents and discard parts with less than 100 examples. Then, we take First-3 as the baseline, which concatenates the top-3 sentences of each source document as the summary.

\begin{figure}
  \centerline{
    \includegraphics[width=0.8\linewidth]{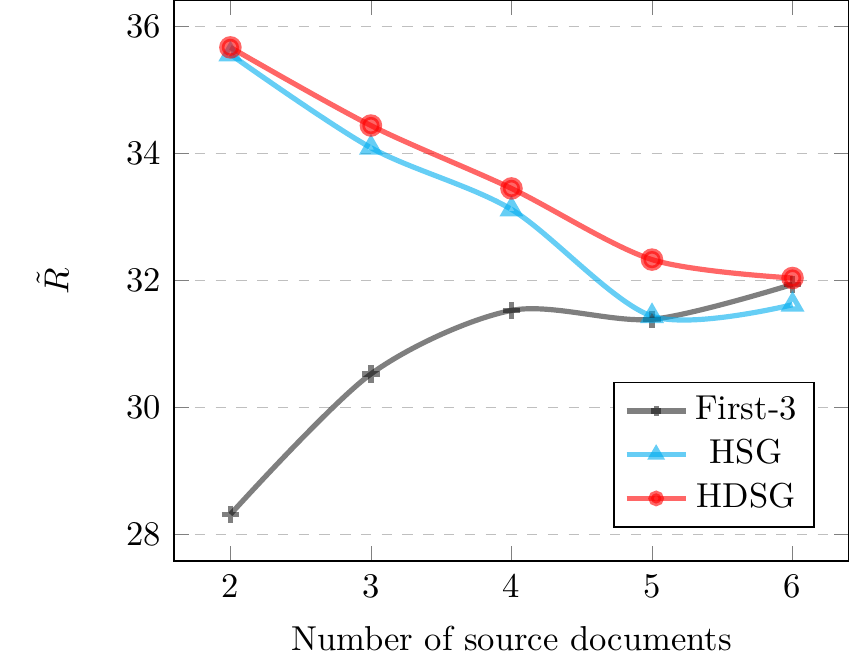}
    }
  \caption{Relationship between number of source documents (x-axis) and $\Tilde{R}$ (y-axis).}
  \label{fig:srcnum}
\end{figure}

In Figure \ref{fig:srcnum}, we can observe that the lead baseline raises while both of our model performance degrade and finally they converge to the baseline. This is because it is more challenging for models to extract limited-number sentences that can cover the main idea of all source documents with the increasing  number of documents. However, the First-3 baseline is forced to take sentences from each document which can ensure the coverage.
Besides, the increase of document number enlarges the performance gap between \textsc{HeterSUMGraph} and \textsc{HeterDocSUMGraph}. This indicates the benefit of document nodes will become more significant for more complex document-document relationships.

\section{Conclusion}

In this paper, we propose a heterogeneous graph-based neural network for extractive summarization. The introduction of more fine-grained semantic units in the summarization graph helps our model to build more complex relationships between sentences
. It is also convenient to adapt our single-document graph to multi-document with document nodes. Furthermore, our models have achieved the best results on CNN/DailyMail compared with non-BERT-based models, and we will take the pre-trained language models into account for better encoding representations of nodes in the future.

\section*{Acknowledgment}
This work was supported by the National Natural Science Foundation of China (No. U1936214 and 61672162), Shanghai Municipal Science and Technology Major Project (No. 2018SHZDZX01) and ZJLab.

\bibliography{acl2020}

\begin{thebibliography}{45}
\expandafter\ifx\csname natexlab\endcsname\relax\def\natexlab#1{#1}\fi

\bibitem[{Arumae and Liu(2018)}]{arumae2018reinforced}
Kristjan Arumae and Fei Liu. 2018.
\newblock Reinforced extractive summarization with question-focused rewards.
\newblock In \emph{Proceedings of ACL 2018, Student Research Workshop}, pages
  105--111.

\bibitem[{Carbonell and Goldstein(1998)}]{carbonell1998use}
Jaime~G Carbonell and Jade Goldstein. 1998.
\newblock The use of mmr, diversity-based reranking for reordering documents
  and producing summaries.
\newblock In \emph{SIGIR}, volume~98, pages 335--336.

\bibitem[{Cheng and Lapata(2016)}]{cheng2016neural}
Jianpeng Cheng and Mirella Lapata. 2016.
\newblock Neural summarization by extracting sentences and words.
\newblock In \emph{Proceedings of the 54th Annual Meeting of the Association
  for Computational Linguistics (Volume 1: Long Papers)}, volume~1, pages
  484--494.

\bibitem[{Devlin et~al.(2018)Devlin, Chang, Lee, and
  Toutanova}]{devlin2018bert}
Jacob Devlin, Ming-Wei Chang, Kenton Lee, and Kristina Toutanova. 2018.
\newblock Bert: Pre-training of deep bidirectional transformers for language
  understanding.
\newblock \emph{arXiv preprint arXiv:1810.04805}.

\bibitem[{Dong et~al.(2018)Dong, Shen, Crawford, van Hoof, and
  Cheung}]{dong2018banditsum}
Yue Dong, Yikang Shen, Eric Crawford, Herke van Hoof, and Jackie Chi~Kit
  Cheung. 2018.
\newblock Banditsum: Extractive summarization as a contextual bandit.
\newblock In \emph{Proceedings of the 2018 Conference on Empirical Methods in
  Natural Language Processing}, pages 3739--3748.

\bibitem[{Durrett et~al.(2016)Durrett, Berg-Kirkpatrick, and
  Klein}]{durrett2016learning}
Greg Durrett, Taylor Berg-Kirkpatrick, and Dan Klein. 2016.
\newblock Learning-based single-document summarization with compression and
  anaphoricity constraints.
\newblock \emph{arXiv preprint arXiv:1603.08887}.

\bibitem[{Erkan and Radev(2004)}]{erkan2004lexrank}
G{\"u}nes Erkan and Dragomir~R Radev. 2004.
\newblock Lexrank: Graph-based lexical centrality as salience in text
  summarization.
\newblock \emph{Journal of artificial intelligence research}, 22:457--479.

\bibitem[{Fabbri et~al.(2019)Fabbri, Li, She, Li, and Radev}]{fabbri2019multi}
Alexander Fabbri, Irene Li, Tianwei She, Suyi Li, and Dragomir Radev. 2019.
\newblock \href {https://doi.org/10.18653/v1/P19-1102} {Multi-news: A
  large-scale multi-document summarization dataset and abstractive hierarchical
  model}.
\newblock In \emph{Proceedings of the 57th Annual Meeting of the Association
  for Computational Linguistics}, pages 1074--1084, Florence, Italy.
  Association for Computational Linguistics.

\bibitem[{Gehrmann et~al.(2018)Gehrmann, Deng, and Rush}]{gehrmann2018bottom}
Sebastian Gehrmann, Yuntian Deng, and Alexander Rush. 2018.
\newblock Bottom-up abstractive summarization.
\newblock In \emph{Proceedings of the 2018 Conference on Empirical Methods in
  Natural Language Processing}, pages 4098--4109.

\bibitem[{Gilmer et~al.(2017)Gilmer, Schoenholz, Riley, Vinyals, and
  Dahl}]{gilmer2017neural}
Justin Gilmer, Samuel~S Schoenholz, Patrick~F Riley, Oriol Vinyals, and
  George~E Dahl. 2017.
\newblock Neural message passing for quantum chemistry.
\newblock In \emph{ICML}, pages 1263--1272.

\bibitem[{Hermann et~al.(2015)Hermann, Kocisky, Grefenstette, Espeholt, Kay,
  Suleyman, and Blunsom}]{hermann2015teaching}
Karl~Moritz Hermann, Tomas Kocisky, Edward Grefenstette, Lasse Espeholt, Will
  Kay, Mustafa Suleyman, and Phil Blunsom. 2015.
\newblock Teaching machines to read and comprehend.
\newblock In \emph{Advances in Neural Information Processing Systems}, pages
  1684--1692.

\bibitem[{Hochreiter and Schmidhuber(1997)}]{hochreiter1997long}
Sepp Hochreiter and J{\"u}rgen Schmidhuber. 1997.
\newblock Long short-term memory.
\newblock \emph{Neural computation}, 9(8):1735--1780.

\bibitem[{Kingma and Ba(2014)}]{kingma2014adam}
Diederik Kingma and Jimmy Ba. 2014.
\newblock Adam: A method for stochastic optimization.
\newblock \emph{arXiv preprint arXiv:1412.6980}.

\bibitem[{Lebanoff et~al.(2018)Lebanoff, Song, and Liu}]{lebanoff2018adapting}
Logan Lebanoff, Kaiqiang Song, and Fei Liu. 2018.
\newblock Adapting the neural encoder-decoder framework from single to
  multi-document summarization.
\newblock In \emph{Proceedings of the 2018 Conference on Empirical Methods in
  Natural Language Processing}, pages 4131--4141.

\bibitem[{LeCun et~al.(1998)LeCun, Bottou, Bengio, Haffner
  et~al.}]{lecun1998gradient}
Yann LeCun, L{\'e}on Bottou, Yoshua Bengio, Patrick Haffner, et~al. 1998.
\newblock Gradient-based learning applied to document recognition.
\newblock \emph{Proceedings of the IEEE}, 86(11):2278--2324.

\bibitem[{Lin and Hovy(2003)}]{lin2003automatic}
Chin-Yew Lin and Eduard Hovy. 2003.
\newblock Automatic evaluation of summaries using n-gram co-occurrence
  statistics.
\newblock In \emph{Proceedings of the 2003 Human Language Technology Conference
  of the North American Chapter of the Association for Computational
  Linguistics}.

\bibitem[{Linmei et~al.(2019)Linmei, Yang, Shi, Ji, and
  Li}]{linmei2019heterogeneous}
Hu~Linmei, Tianchi Yang, Chuan Shi, Houye Ji, and Xiaoli Li. 2019.
\newblock Heterogeneous graph attention networks for semi-supervised short text
  classification.
\newblock In \emph{Proceedings of the 2019 Conference on Empirical Methods in
  Natural Language Processing and the 9th International Joint Conference on
  Natural Language Processing (EMNLP-IJCNLP)}, pages 4823--4832.

\bibitem[{Liu et~al.(2018)Liu, Saleh, Pot, Goodrich, Sepassi, Kaiser, and
  Shazeer}]{liu2018generating}
Peter~J Liu, Mohammad Saleh, Etienne Pot, Ben Goodrich, Ryan Sepassi, Lukasz
  Kaiser, and Noam Shazeer. 2018.
\newblock Generating wikipedia by summarizing long sequences.
\newblock \emph{Proceedings of the 6th International Conference on Learning
  Representations}.

\bibitem[{Liu and Lapata(2019{\natexlab{a}})}]{liu2019hierarchical}
Yang Liu and Mirella Lapata. 2019{\natexlab{a}}.
\newblock Hierarchical transformers for multi-document summarization.
\newblock In \emph{Proceedings of the 57th Annual Meeting of the Association
  for Computational Linguistics}, pages 5070--5081.

\bibitem[{Liu and Lapata(2019{\natexlab{b}})}]{liu2019text}
Yang Liu and Mirella Lapata. 2019{\natexlab{b}}.
\newblock \href {https://doi.org/10.18653/v1/D19-1387} {Text summarization with
  pretrained encoders}.
\newblock In \emph{Proceedings of the 2019 Conference on Empirical Methods in
  Natural Language Processing and the 9th International Joint Conference on
  Natural Language Processing (EMNLP-IJCNLP)}, pages 3721--3731, Hong Kong,
  China. Association for Computational Linguistics.

\bibitem[{Liu et~al.(2019)Liu, Titov, and Lapata}]{liu2019single}
Yang Liu, Ivan Titov, and Mirella Lapata. 2019.
\newblock Single document summarization as tree induction.
\newblock In \emph{Proceedings of the 2019 Conference of the North American
  Chapter of the Association for Computational Linguistics: Human Language
  Technologies, Volume 1 (Long and Short Papers)}, pages 1745--1755.

\bibitem[{Luo et~al.(2019)Luo, Ao, Song, Pan, Yang, and He}]{luo2019reading}
Ling Luo, Xiang Ao, Yan Song, Feiyang Pan, Min Yang, and Qing He. 2019.
\newblock Reading like her: Human reading inspired extractive summarization.
\newblock In \emph{Proceedings of the 2019 Conference on Empirical Methods in
  Natural Language Processing and the 9th International Joint Conference on
  Natural Language Processing (EMNLP-IJCNLP)}, pages 3024--3034.

\bibitem[{Mihalcea and Tarau(2004)}]{mihalcea2004textrank}
Rada Mihalcea and Paul Tarau. 2004.
\newblock Textrank: Bringing order into text.
\newblock In \emph{Proceedings of the 2004 conference on empirical methods in
  natural language processing}, pages 404--411.

\bibitem[{Nallapati et~al.(2017)Nallapati, Zhai, and
  Zhou}]{nallapati2017summarunner}
Ramesh Nallapati, Feifei Zhai, and Bowen Zhou. 2017.
\newblock Summarunner: A recurrent neural network based sequence model for
  extractive summarization of documents.
\newblock In \emph{Thirty-First AAAI Conference on Artificial Intelligence}.

\bibitem[{Nallapati et~al.(2016)Nallapati, Zhou, dos Santos, glar
  Gul{\c{c}}ehre, and Xiang}]{nallapati2016abstractive}
Ramesh Nallapati, Bowen Zhou, Cicero dos Santos, {\c{C}}a~glar Gul{\c{c}}ehre,
  and Bing Xiang. 2016.
\newblock Abstractive text summarization using sequence-to-sequence rnns and
  beyond.
\newblock \emph{CoNLL 2016}, page 280.

\bibitem[{Narayan et~al.(2018)Narayan, Cohen, and Lapata}]{narayan2018ranking}
Shashi Narayan, Shay~B Cohen, and Mirella Lapata. 2018.
\newblock Ranking sentences for extractive summarization with reinforcement
  learning.
\newblock In \emph{Proceedings of the 2018 Conference of the North American
  Chapter of the Association for Computational Linguistics: Human Language
  Technologies, Volume 1 (Long Papers)}, volume~1, pages 1747--1759.

\bibitem[{Paulus et~al.(2017)Paulus, Xiong, and Socher}]{paulus2017deep}
Romain Paulus, Caiming Xiong, and Richard Socher. 2017.
\newblock A deep reinforced model for abstractive summarization.
\newblock \emph{arXiv preprint arXiv:1705.04304}.

\bibitem[{Pennington et~al.(2014)Pennington, Socher, and
  Manning}]{pennington2014glove}
Jeffrey Pennington, Richard Socher, and Christopher Manning. 2014.
\newblock Glove: Global vectors for word representation.
\newblock In \emph{Proceedings of the 2014 conference on empirical methods in
  natural language processing (EMNLP)}, pages 1532--1543.

\bibitem[{Sandhaus(2008)}]{sandhaus2008new}
Evan Sandhaus. 2008.
\newblock The new york times annotated corpus.
\newblock \emph{Linguistic Data Consortium, Philadelphia}, 6(12):e26752.

\bibitem[{See et~al.(2017)See, Liu, and Manning}]{see2017get}
Abigail See, Peter~J Liu, and Christopher~D Manning. 2017.
\newblock Get to the point: Summarization with pointer-generator networks.
\newblock In \emph{Proceedings of the 55th Annual Meeting of the Association
  for Computational Linguistics (Volume 1: Long Papers)}, volume~1, pages
  1073--1083.

\bibitem[{Shi et~al.(2016)Shi, Li, Zhang, Sun, and Philip}]{shi2016survey}
Chuan Shi, Yitong Li, Jiawei Zhang, Yizhou Sun, and S~Yu Philip. 2016.
\newblock A survey of heterogeneous information network analysis.
\newblock \emph{IEEE Transactions on Knowledge and Data Engineering},
  29(1):17--37.

\bibitem[{Tu et~al.(2019)Tu, Wang, Huang, Tang, He, and Zhou}]{tu2019multi}
Ming Tu, Guangtao Wang, Jing Huang, Yun Tang, Xiaodong He, and Bowen Zhou.
  2019.
\newblock Multi-hop reading comprehension across multiple documents by
  reasoning over heterogeneous graphs.
\newblock \emph{arXiv preprint arXiv:1905.07374}.

\bibitem[{Vaswani et~al.(2017)Vaswani, Shazeer, Parmar, Uszkoreit, Jones,
  Gomez, Kaiser, and Polosukhin}]{vaswani2017attention}
Ashish Vaswani, Noam Shazeer, Niki Parmar, Jakob Uszkoreit, Llion Jones,
  Aidan~N Gomez, {\L}ukasz Kaiser, and Illia Polosukhin. 2017.
\newblock Attention is all you need.
\newblock In \emph{Advances in Neural Information Processing Systems}, pages
  5998--6008.

\bibitem[{Velickovic et~al.(2017)Velickovic, Cucurull, Casanova, Romero, Lio,
  and Bengio}]{velickovic2017graph}
Petar Velickovic, Guillem Cucurull, Arantxa Casanova, Adriana Romero, Pietro
  Lio, and Yoshua Bengio. 2017.
\newblock Graph attention networks.
\newblock \emph{arXiv preprint arXiv:1710.10903}.

\bibitem[{Wang et~al.(2019{\natexlab{a}})Wang, Liu, Zhong, Fu, Qiu, and
  Huang}]{wang2019exploring}
Danqing Wang, Pengfei Liu, Ming Zhong, Jie Fu, Xipeng Qiu, and Xuanjing Huang.
  2019{\natexlab{a}}.
\newblock Exploring domain shift in extractive text summarization.
\newblock \emph{arXiv preprint arXiv:1908.11664}.

\bibitem[{Wang et~al.(2019{\natexlab{b}})Wang, Chang, and Huang}]{wang2019user}
Hsiu-Yi Wang, Jia-Wei Chang, and Jen-Wei Huang. 2019{\natexlab{b}}.
\newblock User intention-based document summarization on heterogeneous sentence
  networks.
\newblock In \emph{International Conference on Database Systems for Advanced
  Applications}, pages 572--587. Springer.

\bibitem[{Wei(2012)}]{wei2012document}
Yang Wei. 2012.
\newblock Document summarization method based on heterogeneous graph.
\newblock In \emph{2012 9th International Conference on Fuzzy Systems and
  Knowledge Discovery}, pages 1285--1289. IEEE.

\bibitem[{Xu and Durrett(2019)}]{xu2019neural}
Jiacheng Xu and Greg Durrett. 2019.
\newblock Neural extractive text summarization with syntactic compression.
\newblock \emph{arXiv preprint arXiv:1902.00863}.

\bibitem[{Xu et~al.(2019)Xu, Gan, Cheng, and Liu}]{xu2019discourse}
Jiacheng Xu, Zhe Gan, Yu~Cheng, and Jingjing Liu. 2019.
\newblock Discourse-aware neural extractive model for text summarization.
\newblock \emph{arXiv preprint arXiv:1910.14142}.

\bibitem[{Yasunaga et~al.(2017)Yasunaga, Zhang, Meelu, Pareek, Srinivasan, and
  Radev}]{yasunaga2017graph}
Michihiro Yasunaga, Rui Zhang, Kshitijh Meelu, Ayush Pareek, Krishnan
  Srinivasan, and Dragomir Radev. 2017.
\newblock Graph-based neural multi-document summarization.
\newblock \emph{arXiv preprint arXiv:1706.06681}.

\bibitem[{Zhang et~al.(2018)Zhang, Lapata, Wei, and Zhou}]{zhang2018neural}
Xingxing Zhang, Mirella Lapata, Furu Wei, and Ming Zhou. 2018.
\newblock Neural latent extractive document summarization.
\newblock In \emph{Proceedings of the 2018 Conference on Empirical Methods in
  Natural Language Processing}, pages 779--784.

\bibitem[{Zhong et~al.(2020)Zhong, Liu, Chen, Wang, Qiu, and
  Huang}]{zhong2020extractive}
Ming Zhong, Pengfei Liu, Yiran Chen, Danqing Wang, Xipeng Qiu, and Xuan-Jing
  Huang. 2020.
\newblock Extractive summarization as text matching.
\newblock In \emph{Proceedings of the 58th Conference of the Association for
  Computational Linguistics}.

\bibitem[{Zhong et~al.(2019{\natexlab{a}})Zhong, Liu, Wang, Qiu, and
  Huang}]{zhong2019searching}
Ming Zhong, Pengfei Liu, Danqing Wang, Xipeng Qiu, and Xuan-Jing Huang.
  2019{\natexlab{a}}.
\newblock Searching for effective neural extractive summarization: What works
  and what’s next.
\newblock In \emph{Proceedings of the 57th Annual Meeting of the Association
  for Computational Linguistics}, pages 1049--1058.

\bibitem[{Zhong et~al.(2019{\natexlab{b}})Zhong, Wang, Liu, Qiu, and
  Huang}]{zhong2019closer}
Ming Zhong, Danqing Wang, Pengfei Liu, Xipeng Qiu, and Xuan-Jing Huang.
  2019{\natexlab{b}}.
\newblock A closer look at data bias in neural extractive summarization models.
\newblock In \emph{Proceedings of the 2nd Workshop on New Frontiers in
  Summarization}, pages 80--89.

\bibitem[{Zhou et~al.(2018)Zhou, Yang, Wei, Huang, Zhou, and
  Zhao}]{zhou2018neural}
Qingyu Zhou, Nan Yang, Furu Wei, Shaohan Huang, Ming Zhou, and Tiejun Zhao.
  2018.
\newblock Neural document summarization by jointly learning to score and select
  sentences.
\newblock In \emph{Proceedings of the 56th Annual Meeting of the Association
  for Computational Linguistics (Volume 1: Long Papers)}, volume~1, pages
  654--663.

\end{thebibliography}
\bibliographystyle{acl_natbib}

\appendix

\section{Appendices}
\label{sec:appendix}
In order to select the best iteration number for \textsc{HeterSUMGraph}, we compare performances of different $t$ on the validation set of CNN/DM. All models are trained on a single GeForce RTX 2080 Ti GPU for about 5 epochs. As Table \ref{tab:Iteration} shows, our \textsc{HeterSUMGraph} has comparable results for $t=1$ and $t=3$. However, when the iteration number goes from 1 to 3, the time for one epoch nearly doubles. Therefore, we take $t=1$ as a result of the balance of time cost and model performance.

\begin{table}[htbp] \small
  \centering \renewcommand\arraystretch{1}
    \begin{tabular}{lrrrr}
    \toprule
    \textbf{Number} & \multicolumn{1}{c}{\textbf{R-1}} & \multicolumn{1}{c}{\textbf{R-2}} & \multicolumn{1}{c}{\textbf{R-L}} & \multicolumn{1}{c}{\textbf{Time}} \\
    \midrule
    $t$ = 0     & 43.63 & 19.58 & 37.39 & 3.16h \\
    $t$ = 1      & 44.26 & 19.97 & 38.03 & 5.04h \\
    $t$ = 2      & 44.13 & 19.85 & 37.87 & 7.20h \\
    $t$ = 3      & 44.28 & 19.96 & 37.98 & 8.93h \\
    \bottomrule
    \end{tabular}%
    \caption{Different turns of iterative updating of sentence nodes. The experiments are performed on the validation set of CNN/DM. Time is the average time of one epoch.}
  \label{tab:Iteration}%
\end{table}%


\end{document}